%% file: main.tex
\documentclass[letterpaper, 10 pt, conference]{ieeeconf}  
\usepackage{graphicx}
\usepackage{amsmath}
\usepackage{amssymb}
\usepackage{placeins}
\usepackage{subcaption}
\usepackage{caption}
\usepackage{xcolor}
\usepackage{hyperref}
\usepackage{booktabs}
\usepackage{cite}
\usepackage{multirow}

\usepackage{tcolorbox}
\newcommand{\minus}{\scalebox{0.8}{$-$}}
\newcommand{\plus}{\scalebox{0.6}{$+$}}
\usepackage{siunitx}
\usepackage{xcolor}
\usepackage{amsmath}
\usepackage{amssymb}
\usepackage{algorithm,algpseudocode}
\usepackage{pifont}
\usepackage{url}

\newcommand{\optimalpi}{\pi^*}

\newcommand{\nop}[1]{}

\DeclareMathOperator*{\argmin}{arg\,min}

\IEEEoverridecommandlockouts                              

\title{\LARGE \bf
Learning to Control DC Motor for Micromobility in Real Time with Reinforcement Learning
}

\author{Bibek Poudel$^{*1}$, Thomas Watson$^{*2}$, Weizi Li$^{1}$
\thanks{*equal contribution}
\thanks{$^{1}$Department of Computer Science, University of Memphis,
        Memphis, TN 38152, USA 
        {\tt\small \{bpoudel,wli\}@memphis.edu}}%
\thanks{$^{2}$Department of Electrical and Computer Engineering, University of Memphis,
        Memphis, TN 38152, USA
        {\tt\small tpwatson@memphis.edu}}
}

\begin{document}

\maketitle

\begin{abstract}

Autonomous micromobility has been attracting the attention of researchers and practitioners in recent years. A key component of many micro-transport vehicles is the DC motor, a complex dynamical system that is continuous and non-linear. Learning to quickly control the DC motor in the presence of disturbances and uncertainties is desired for various applications that require robustness and stability. Techniques to accomplish this task usually rely on a mathematical system model, which is often insufficient to anticipate the effects of time-varying and interrelated sources of non-linearities. While some model-free approaches have been successful at the task, they rely on massive interactions with the system and are trained in specialized hardware in order to fit a highly parameterized controller. In this work, we learn to steer a DC motor via sample-efficient reinforcement learning. Using data collected from hardware interactions in the real world, we additionally build a simulator to experiment with a wide range of parameters and learning strategies. With the best parameters found, we  learn an effective control policy in one minute and 53 seconds on a simulation and in 10 minutes and 35 seconds on a physical system.

\end{abstract}

\input{sections/intro}
\input{sections/related}
\input{sections/prelim}

\input{sections/method}

\input{sections/results}

\input{sections/conclusion}


\bibliographystyle{IEEEtran}
\bibliography{ref}

\end{document}

%% file: sections/intro.tex
\section{Introduction}
\label{sec:intro}

While significant progress has been made in developing full-size autonomous vehicles, autonomous micromobility, including golf carts~\cite{cart2015}, scooters~\cite{scooter1,scooter2}, and bikes~\cite{bike1,Lin2019BikeTRB}, has also seen a rise~\cite{micro2021}. A key component to enable these micro-transport vehicles is the Direct Current (DC) motor. In this work, we focus on the control of DC motors for steering via reinforcement learning (RL) in real time. 

RL has seen many successes over the past decade on various applications such as games~\cite{mnih2013playing, silver2016mastering}, robotics~\cite{levine2016end}, natural language processing~\cite{bahdanau2016actor}, and intelligent transportation systems~\cite{el2013multiagent,wu2021flow}.
However, in order to learn an effective policy, most RL applications to date require an offline virtual environment for collecting a large number of experiences much faster than real-time. 
This phenomenon is pronounced when we combine the use of deep neural networks and RL since insufficient data points will not fit a deep neural network properly and enable it to generalize well. 
The learning task is further complicated if a physical system is involved: 1) the system may contain a continuous state space; 2) we may lack a precise mathematical model to describe the system dynamics; 3) the system is likely to experience both internal uncertainties (noise, inertia, control oscillation) and external perturbations (friction, timing restraints); 
and 4) due to the reasons 1)--3), the success of policy transfer from the virtual environment to real-world is not guaranteed.  

In this work, we experiment with the original idea of RL, where an effective policy is learned, tested, and improved in real-time on a physical system. The system dynamics are continuous and non-linear. No knowledge is assumed of the system internal factors and external environment perturbations. Specifically, we consider position/angle control of a DC motor attached to a golf cart steering wheel: given a target steering position, we learn how to apply a sequence of voltages to reach that position as fast as possible.
The system dynamics are non-linear due to many factors, including temperature's influence on the resistance and impedance of the motor coil, electromagnetic effects (cogging torque, eddy current), and the presence of dead-zones (minimum voltage required in either polarity to rotate the motor) caused by friction and variable load~\cite{buechner2013nonlinear, kara2004nonlinear}. The difficulty in mathematically modeling the co-occurrence of these time-varying and interdependent factors significantly complicates the control task. 

Traditional position control approaches for a DC motor range from a simple Bang-bang controller~\cite{bellman1956bang} to more sophisticated methods such as Proportional Integral Derivative (PID) controller~\cite{aylor1980design} and Model Predictive Controller (MPC)~\cite{qin2003survey}. To reach the desired position, a Bang-bang controller 
applies a maximum voltage in one polarity until a switching point, then applies maximum voltage in the opposite polarity. However, during this process, the motor will experience oscillation and overshoot errors due to inertia, thus fails the task. On the other hand, a PID controller requires an expert to tune control parameters, which could be labor-intensive. MPC, which can anticipate future events before exerting a control signal, has emerged as a substitute of the PID controller (which lacks predictive capabilities). However, MPC requires a dynamic model of the system, which is unavailable to us. Recently, supervised learning with neural networks has been proposed to solve control tasks~\cite{ross2011reduction}, yet they usually rely on a static dataset from limited explorations of the state space,
hence do not encompass a wide range of sources of non-linearities and uncertainties that present in our physical system. 

\begin{figure*}[t]
	\includegraphics[width=\textwidth]{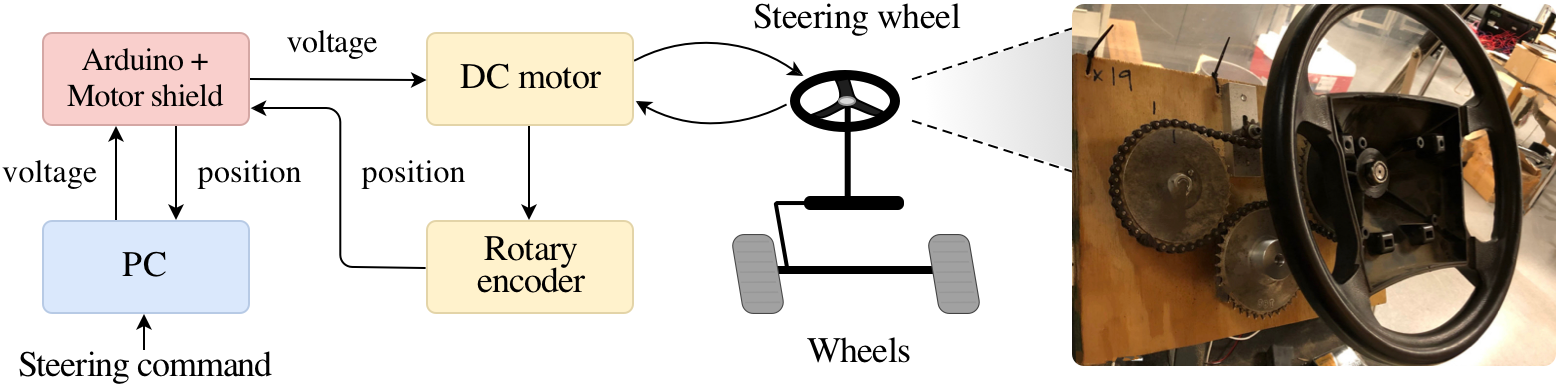}
	\caption{\small{Schematic diagram of our control task. The steering commands are input to a PC, which are then converted to voltage signals (interfaced through an Arduino) and applied to the motor. The feedback signals to the PC are obtained from the rotary encoder.}}
	\label{fig:schematic}
 	\vspace{-1.5em}
\end{figure*}

As an alternative, model-free RL~\cite{sutton2018reinforcement}, a data-driven approach that aims to learn a control policy based directly on reward feedback from the controller's interactions with the system, is more promising for our task. One such algorithm is Neural fitted Q (NFQ)~\cite{riedmiller2005neural}, where a state-action Q-value estimator is iteratively improved using transitions collected from the physical system itself. NFQ has been used to learn control policies for a cart-pole system~\cite{riedmiller2005neural}, a soccer playing robot~\cite{hafner2007neural}, and a real car~\cite{riedmiller2007learning}. However, learning to automatically control a DC motor position in real-time without system knowledge or manual interference is a challenging control task itself. 
In this work, the approach to learning a control strategy is data-driven and can be applied to learn a control policy in any environment, i.e., under specific uncertainties and non-linearities of numerous applications of DC motors in industrial, household, and robotic processes.


In addition to real-world experiments, we build a simulator using the collected data from the physical system. The purpose of simulations is to 1) prevent damage to the hardware from acceleration and jerk induced by an imperfect policy; and 2) experiment with a wide range of hyperparameters and learning strategies associated with NFQ---a study that is largely omitted from previous literature. Furthermore, operating a golf cart in the real world could be physically dangerous under a sub-optimal policy and amongst people. Although the simulated environment is built from real-world data, transitions made in the simulation are not equivalent to retrieving them from stored memory: every episode starts from a randomly initialized position (hence a randomly initialized state) which may not always have an exact match in simulation due to approximation errors.     

As a result, starting without prior knowledge and relying solely on real-world interactions, we successfully learned a control policy in $10$ minutes and $35$ seconds in hardware. Some of the strategies and parameters used to learn in hardware are determined from a simulator. In addition, using the results of experiments in the simulator (number of neural network parameters $=91$, hint-to-goal size $=2\%$, linearly decaying exploration strategy, neural network reset frequency $=$~every~$50$ episodes, and exponentially increased curriculum learning for position initialization), we also successfully learned a control policy in simulation under one minute and $53$ seconds (on average) using an off-the-shelf laptop. 

%% file: sections/related.tex
\section{Related Work}
\label{sec:related}

DC motors offer a vast and adjustable operating range of speed and position, resulting in their extensive use in robotics and industrial control processes such as robotic arm and conveyor~\cite{horng1999neural}. 
Since their invention in the mid-nineteenth century, numerous position control approaches have emerged among which a Proportional-Integral-Derivative (PID) controller~\cite{meshram2012tuning} is commonly adopted. A PID controller consists of a circuit or a program that corrects an observed error (negative feedback) by exercising a control signal to oppose it, stabilizing any external influences. Control parameters (gain, integral time, and derivative time), usually tuned by an experienced operator using ``trial and error'', determine the intensity of the error correction. Therefore, a reliable operation of the controller depends on the quality of parameter tuning. Although several attempts exist in making PID controllers operate reliably on DC motor control, e.g., combining the parameter tuning with a genetic algorithm~\cite{thomas2009position} or RL~\cite{khater2015embedded}, they still require and rely on an underlying model to describe system behaviors.

In the last three decades, Model Predictive Controller (MPC) has emerged as a significant advancement in control theory, effectively replacing PID controllers in a majority of complex control problems~\cite{alkurawy2018model, syaichu2011model}. MPC relies on the knowledge of the system in the form of its expected dynamic behavior (model) over a prediction horizon and a loss function that describes desired control behaviors (e.g., track a reference trajectory by reducing aggressiveness). Among other variables, the loss function consists of current and future inputs. Using this formulation, once the optimal sequence of inputs for future timesteps is determined, only the input to the current timestep is applied to the system, and the future inputs are iteratively computed in a reduced horizon~\cite{sahoo2015optimal}. Since the predictions do not account for future feedback, the optimal input at every timestep inherently does not account for uncertainties but solely based on the system's model~\cite{lee2011model}. Moreover, to design a loss function that simultaneously satisfies multiple control objectives (which can often contradict each other) is challenging~\cite{vazquez2016model}. Recently, supervised learning has demonstrated better convergence and generalization properties in either a stand-alone manner~\cite{aamir2013replacing} or combined~\cite{horng1999neural} with traditional techniques for DC motor control. However, it usually operates on a static dataset which could be unrepresentative of a real-world environment and its time-varying and stochastic features.

Systems that use RL for control and consist of a DC motor (but not explicitly controlling DC motor position or speed) have been examined~\cite{riedmiller2005neural, caarls2015parallel}. Nevertheless, RL is relatively new to the explicit position or speed control of electric drives such as a DC motor.~\cite{schenke2019controller} establish a proof of concept where they introduce deep RL algorithms (value-based, policy-based, actor-critic) for control of electric drives
in simulation. Their work is followed by~\cite{traue2020toward}, who develop a toolbox to design, train, and benchmark traditional and RL-based control strategies in simulation. However, at the time of this writing, the toolbox does not have hardware interfacing capabilities. Recently,~\cite{book2021transferring}
successfully transferred a controller learned in simulation to a real-world electric drive; the control policy is learned using deep RL algorithms thus still requires an offline virtual environment to generate training data. 

In this work, we adopt Neural Fitted Q~\cite{riedmiller2005neural} where fitted value iteration is performed in a neural network with four layers ($61$ parameters in hardware, $91$ parameters in simulation) using real-world interaction data. The learning setup enables the agent to explore hardware features and acquire real-time experiences to learn an effective policy.


%% file: sections/prelim.tex
\section{Preliminaries}
\label{sec:prelim}
In this section, we first introduce our problem formulation using RL, along with the Q-learning algorithm~\cite{watkins1989learning} and its deep learning variants. Then, we present the NFQ algorithm~\cite{riedmiller2005neural} and briefly discuss its use on real-time control of a physical system.

\subsection{Reinforcement Learning}
We model our problem as a Markov Decision Process (MDP) which is described by a tuple $<\mathcal{S}, \mathcal{A}, P, R,\gamma>$, where $S$ is the state space, $A$ is the action space, $P(s,a,s')$ is the stochastic transition function, $R$ is the immediate reward $R:\mathcal{S} \times \mathcal{A} \rightarrow \mathbb{R}$, and $\gamma\in[0,1]$ is the discount factor. 
In an episodic task of $T$ timesteps, the goal of the agent is to maximize the cumulative discounted reward (return) $G_t=\sum_{k=t+1}^{T} \gamma^{k-t-1}R_{k}$ by selecting actions using its policy $\pi: \mathcal{S} \rightarrow \mathcal{A}$. The state-action value function $Q_\pi (s, a) = \mathbb{E}_\pi [G_t|S_t = s, A_t = a]$ gives an estimate of the expected cumulative reward obtained by taking an action $a \in \mathcal{A}$ at time $t \in [1,T]$ in state $s \in \mathcal{S}$ and following the policy $\pi$ thereafter. In classical Q-learning, the Q-function is iteratively updated:   
\begin{equation}
Q_{(s, a)}=\left(1-\alpha\right)Q_{\left(s, a\right)} + \alpha [R_{\left(s,a\right)} + \gamma \max_{a'} Q_{(s', a')}], 
\end{equation} 
\noindent where $\alpha$ denotes the learning rate and $a'$ is the successor action of $a$. For a large state space (e.g., continuous), it is intractable to maintain Q values in a tabular representation; hence approximation techniques are required, among which neural networks are a popular choice due to their generalization and convergence properties. As an example, Deep Q-Network (DQN)~\cite{mnih2013playing} enhances Q-learning using a deep neural network as a function approximator, and uses Experience Replay and Fixed Q-Targets to improve its performance. 
\nop{
Among its contributions are experience replay (store observed transitions in memory and sample them uniformly at training) and target network (additional Q-value estimator). Experience replay reduces variance and the effects of non-stationary data distribution caused by policy updates because it allows learned policy to not just rely on current experiences. On the other hand, the policy being learned minimizes the error between Q-value estimates of the target network and its own estimates. While the learned policy is updated every iteration, the parameters of the target network are copied from the learned policy every $\tau$ timesteps and fixed all other timesteps; this was shown to stabilize the algorithm, i.e., prevent policy divergence. As a result, DQN achieved state-of-the-art performance in multiple Atari 2600 games while just relying on high-dimensional raw pixels as inputs. 
}
Later improvements include DQN with Prioritized Experience Replay~\cite{schaul2015prioritized}, which increases the sampling probability of experiences that have higher expected return; Double DQN~\cite{van2016deep} which prevents over-optimistic value estimates in DQN by using separate estimators for action selection and evaluation; and Dueling DQN~\cite{wang2016dueling} which assumes that actions may not meaningfully affect the environment every timestep and hence updates the estimators of state and action advantage values separately.



\subsection{Neural Fitted Q (NFQ)}
Although deep Q-learning is effective in many tasks, they require massive data (e.g., millions of trajectories) and intensive computing resources (e.g., high-end Graphics Processing Units), due to the requirement of fitting deep neural networks appropriately. These requirements limit deep Q-learning to be used in real-time learning and control of a physical system.

NFQ, on the other hand, is a data-efficient RL algorithm (shown in Algorithm~\ref{alg:nfq}). NFQ belongs to a class of model-free RL algorithms known as fitted value iteration, where value iteration is performed on data collected online, in real-time. In NFQ, the function approximator is a Multilayer Perceptron (MLP) which, in every iteration, is trained using all transitions the system has experienced so far. The reuse of stored transitions not only enables NFQ attain data efficiency and stability in learning, but also allows batch-mode operation where supervised learning can be applied. The algorithm consists of two major steps: 1) the generation of training data and 2) training MLPs at every iteration using the $R_{prop}$ optimizer. In terms of applications, NFQ has been used to learn effective control policies for a cart-pole system~\cite{riedmiller2005neural}, a soccer playing robot~\cite{hafner2007neural}, and a real car~\cite{riedmiller2007learning}.

\begin{algorithm}
\caption{Neural Fitted Q}
\begin{algorithmic}[1]
\State \textbf{Input:} $D$ transition samples
\State $Q_{0}~\gets$~Initialize MLPs
\For{iteration k $\in$ [1, N]}
    \State Generate pattern set,
    \State $\mathbf{P}=\{(input^l, target^l), l = 1,2,...~D\}$, where:
    \State\hspace*{1em}$input^l = s^l, a^l$
    \State\hspace*{1em}$target^l = c(s^l, a^l) + \gamma \min_b Q_k(s'^l,b)$
    \State $Q_{k+1}~\gets~$Train on $\mathbf{P}$ \hspace*{0.5em}//~\textit{via supervised learning}
    \State $k~\gets~k + 1$
\EndFor
\end{algorithmic}
\label{alg:nfq}
\end{algorithm}

\begin{figure*}[h!]
	\centering
	\includegraphics[width=\textwidth]{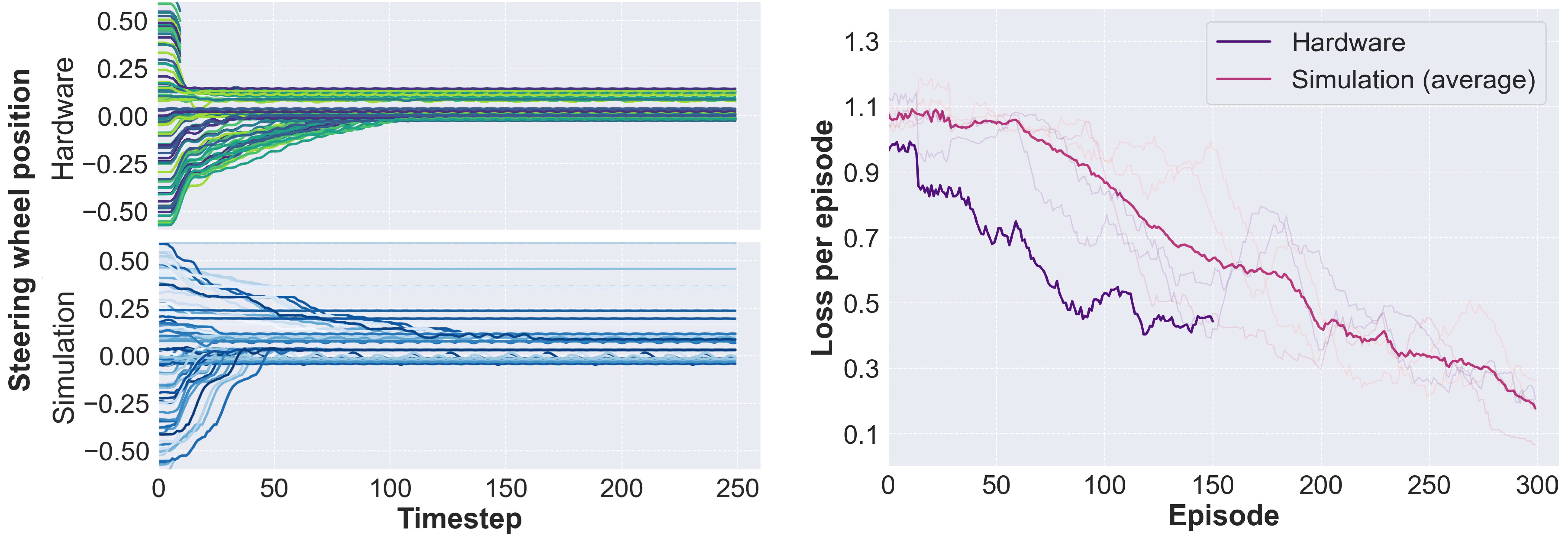}
 	\vspace{-1em}
	\caption{\small{LEFT: Position trajectories with the same $100$ arbitrarily initializations in hardware and simulation where $\pm0.05$ is the success range. $21$ trajectories in the hardware initialized between $0.25$ and $0.5$ fail due to high velocity, while trajectories initialized within $\pm0.25$ have a higher chance of success. Due to approximation errors in simulation, $21$ trajectories ``stall'' in their initialized states, i.e., neither fail nor succeed until the end of $250$ timesteps. RIGHT: The results of our control task in simulation and on hardware (partially using best parameters and settings determined from the simulation, e.g., size of hint-to-goal $=2\%$, network reset frequency$=50$). In the simulation, the results are averaged over five randomized neural network weights initializations; the training loss per episode on average decreases after the first reset at $50$ episodes to reach the vicinity of $0.2$ at episode $300$, indicating a high probability of successful trajectory completion, while on hardware the loss per episode converges to $0.4$. In both cases, the training loss per episode is shown as a moving average with a window size of $30$. On average, the training time in simulation for $300$ episodes is one minute and $53$ seconds in a Macbook Air with an M1 processor and $8$GB RAM. In contrast, training on hardware with real-world interactions for $150$ episodes is complete in $10$ minutes and $35$ seconds in a Macbook Pro with an intel core-i7 processor and $16$GB RAM.}} 
	\label{fig:simulation}
 	\vspace{-1.5em}
\end{figure*}

%% file: sections/method.tex
\section{Methodology}
We consider a $T$-step control task. Denoting the actual cost of performing action $a$ in state $s$ as $C\left(s,a\right)$, we aim to find a policy within the policy class $\Pi$ to minimize the cumulative cost over $T$ steps:
\begin{equation}
\optimalpi = \argmin_{\pi \in \Pi} \sum_{t=1}^{T} C\left(s_{t}, a_{t} \sim \pi\left(s_t\right)\right).
\label{eq:true}
\end{equation}

\noindent Eq.~\ref{eq:true} can be challenging to solve because $C$ is difficult to measure due to all the factors that could affect the physical system (e.g., temperature, electromagnetic effects, etc) and the environment state $s$ is usually not fully observed.  

Given the difficulties and objective, instead of minimizing $C$, it is more practical to minimize the observed surrogate loss $L(\phi, a)$, where $\phi=\phi(s)$ is the (partial) observation of state $s$, $a = \pi(\phi)$ is the learner's action given $\phi$. We further denote the distribution of observations encountered by $\pi$ at $t$ as $ \mathcal{D}_{\pi,t} $, which is the result of executing $ \pi $ from step $ 1 $ to $ t-1 $. Then, $\mathcal{D}_{\pi}=\frac{1}{T}\sum_{t=1}^{T}\mathcal{D}_{\pi,t}$ is the averaged distribution of observations over $ T $ steps, induced by $\pi$. Our goal is then to obtain a policy that can minimize the observed surrogate loss in its own induced observation distribution:

\begin{equation}
\optimalpi = \argmin_{\pi \in \Pi} \mathbb{E}_{\phi \sim \mathcal{D}_{\pi}, a \sim \pi(\phi)} \left[L\left(\phi,a\right)\right].
\label{eq:surrogate}
\end{equation}

\noindent Because $\mathcal{D}_\pi$ can be sampled only by executing $\pi$ in the environment, and the actions generated by $\pi$  and the observations encountered by $\pi$ are intertwined, Eq.~\ref{eq:surrogate} represents a non-convex optimization problem (even $L$ can be convex) under non-i.i.d. conditions. Several approaches exist to Eq.~\ref{eq:surrogate}, among which one option is supervised learning. However, supervised learning trains on a static dataset which may not sufficiently covering the state space that consists of many sources of non-linearities and uncertainties of a physical system. In addition, supervised learning works by assuming the underlying data is i.i.d.; applying it to a non-i.i.d. task will result in $O(\epsilon T^2)$ learning error rather than its typical error $O(\epsilon T)$~\cite{ross2011reduction}. This could be problematic for many control tasks, especially when a physical system is involved. 

\begin{figure*}[h]
	\includegraphics[width=\textwidth]{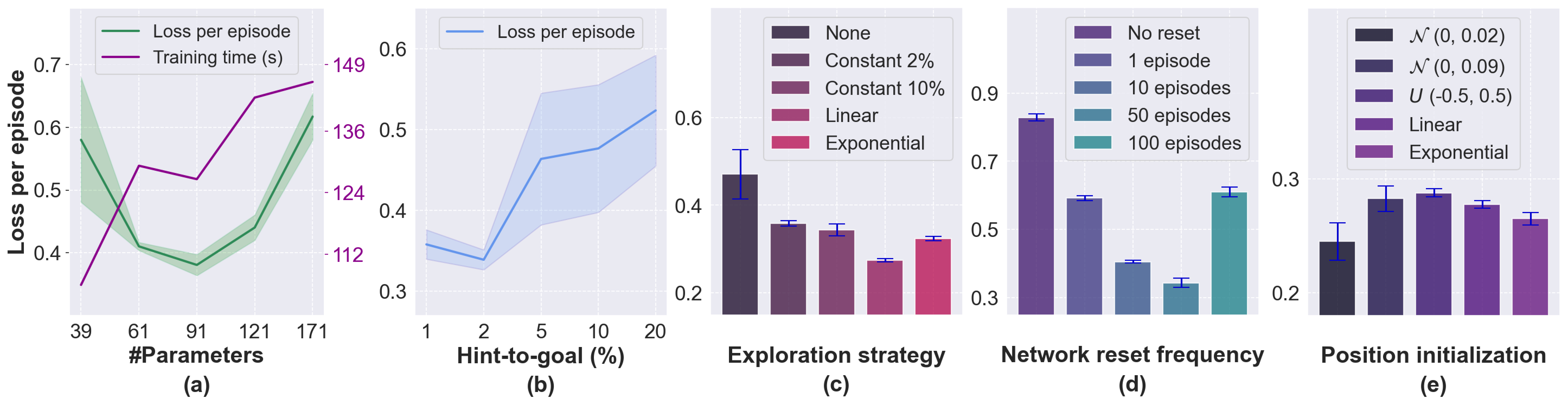}
	\vspace{-1.5em}
	\caption{\small{Experiments on various parameterizations and learning strategies in simulation. Each result is averaged over five different random neural network weight initializations during training and $10$ episodes of testing. In each initialization, the NFQ algorithm is trained for $290$ episodes and tested for $10$ episodes. All experiments (training and test) are performed on a Macbook Air with an M1 processor and $8$GB RAM. a) Average loss per episode decreases (increasing the probability of success) as the number of parameters increases from $39$ to $91$; further increasing of the number of parameters incurs higher losses. b) The size of artificially induced transitions in training set at $2\%$ has the lowest loss. c) Among various exploration strategies, linearly decaying the probability of exploration incurs the least loss. d) Resetting the weights of the neural network every $50$ episodes converges to a lower training loss. e) sampling the steering wheel position initializations closer to the goal position $(0)$ from a Gaussian distribution $\mathcal{N}(0, 0.02)$ leads to the lowest loss.}}
	\label{fig:sim_experiments}
	\vspace{-1.5em}
\end{figure*}

Another approach to Eq.~\ref{eq:surrogate} is RL. To be specific, DC motor position control is a minimum-time control problem~\cite{riedmiller201210}, where the control objective is to reach the desired state in minimum time. In this work, we choose a state definition that sufficiently captures the dynamic behaviors of the system, which is \{current steering wheel position, (angular) velocity, and last applied voltage\}. The action space consists of two voltage commands of equal magnitude and opposite polarity for exerting the torque of the motor in either direction. The action is be made solely on the current state (not the history) with the aim to reach the desired state. This feature satisfies the Markov property and hence our formulation of the control task as an MDP. Note that the actual state of our physical system is not precisely attainable. Instead, we obtain the (partial) observation of a state ($\phi$) which includes the snapshot of the measurements of the motor's position and velocity, and the last applied voltage.  

Specifically, we measure the position of the steering wheel in fractions of a circle, e.g., $0.1=36\si{\degree}$; we measure the velocity as the difference of current and last positions at every timestep; finally, we measure the voltage changes by $\pm0.1 V$ from its last applied value. The actions can be either \textit{turning the motor to left} or \textit{turning the motor to right} as determined by the polarity of applied voltage. The success of the control task is defined by goal states (denoted by ${X}^{\plus}$), which constitute all the states that fulfill the control criteria. The failure (the controller stops) is defined by entering forbidden or undesired states (denoted by ${X}^{\minus}$). The goal and forbidden states are defined as (\textit{SW}$=$ steering wheel):  
\vspace{-0.1em} 
\begin{equation}
\phi \in
\begin{cases} 
    X^{\plus}\text{~if~}-0.05>\textit{SW position}<0.05\text{~and}\\
    \hspace{3em}-0.01>\textit{SW velocity}<0.01, &\\
    X^{\minus}\text{~if~}-0.7>\textit{SW position}<0.7\text{~and}\\
    \hspace{3em}-0.04>\textit{SW velocity}<0.04.\\
\end{cases}
\end{equation}

\noindent The steering wheel position is initialized in the range $[-0.5, 0.5]$, and the velocity is initialized to zero. Our control objective is to reach the goal states as quickly as possible and stay there until the end of the episode (each episode is $250$ timesteps) while avoiding forbidden states. This behavior is incentivized by the observed surrogate loss defined as follows:  
\vspace{-0.1em} 
\begin{equation}
  L(\phi, a) =
    \begin{cases}
      0 & \text{if $\phi~\in~X^+$},\\
      1 & \text{if $\phi~\in~X^-$},\\
      0.001\times~y & \forall~\phi~\not\in~(X^+\cup~X^-),\\
      \end{cases} \\
\end{equation}
where $y=2$ if the steering wheel is moving away from the goal states and $y=1$ otherwise. The doubling of the loss is to discourage transitions towards the forbidden states.


%% file: sections/results.tex
\section{Experiments and Results} 
In this section, we first introduce the results of our physical system. We then introduce the results in simulation of various algorithmic parameterizations and learning strategies.  

\subsection{Hardware Results}
Our physical system is a golf cart modified for electric control of the steering. An electric drill motor is re-purposed to turn a cog, which turns the steering wheel through an attached chain. The chain also drives a rotary encoder so that the angular position of the steering wheel can be tracked. Both motor and rotary encoder are connected to an Arduino, which communicates to a PC. The PC sends commands to the Arduino to set the applied motor voltage, and the Arduino replies with the wheel's position. This communication cycle occurs every 20 milliseconds (50Hz) so that rapid adjustment of the motor voltage and responsive tracking of the position are possible.

Regarding NFQ settings, the MLP's architecture is $4$ layered: $4$ input neurons, $2$ hidden layers each with $5$ neurons, and an output layer with $1$ neuron. In addition, we use several techniques from Riedmiller~\cite{riedmiller201210}: trajectory initialization strategy (sampling the starting steering wheel position uniformly at random in $[-0.55, +0.55]$ to ensure sufficient exploration of the state space), growing batch-size technique (the number of transitions making a training batch grows as learning progresses), and hint-to-goal heuristic (adding artificial transitions with zero loss to each batch to ensure the success of the agent, even at early stages of learning). To balance exploration and exploitation, an $\epsilon$-greedy strategy with $\epsilon=10\%$ is adopted. 
The result of $150$ episodes is shown in Figure~\ref{fig:simulation} LEFT. The loss before the first reset is generally high; the loss in the middle falls but still has spikes, and the loss after the second reset is generally low. Out of the 150 episodes, the number of trajectories that successfully fulfilled the control objective is $24$, with the average loss per episode in the last $10$ steps equal to $0.41$. All $150$ episodes took $31,751$ timesteps, totaling $10$ minutes and $35$ seconds of hardware interactions on a Macbook Pro with an Intel i7 processor and 16GB RAM.


\subsection{Simulation Results}

In addition to train and learn on the physical system, we can use the simulation to experiment with a wide range of learning strategies and hyperparameters without damaging the hardware and the risk of potential accidents due to sub-optimal policies. We build the simulation based on the one-nearest neighbor algorithm to the $121,511$ transitions collected from the hardware. For every state that is encountered in simulation, we first find the nearest neighbor to that state and then we assign the state to which the neighbor transitions as the next state. A $100$ arbitrarily initialized real-world and simulated trajectories are shown in Figure~\ref{fig:simulation} LEFT. The trajectories with starting positions $>0.25$ have a higher chance of failure on both systems. Of the physical system, $21$ such trajectories fail due to high velocity, whereas in simulation, $21$ of them neither fail nor succeed until the end of timesteps.

The experiments conducted on various learning strategies and hyperparameters are shown in Figure~\ref{fig:sim_experiments}. These experiments include \ding{182} varying the parameter count of the neural network at values $39, 61, 91, 121, 171$ to examine loss per episode and total time taken to train for $290$ epochs. We find that with $91$ parameters, the loss per episode is lowest at $0.38$ with a total training time of $126$ seconds. \ding{183} The number of hint-to-goal transitions to be included in the training set at each iteration are varied at five levels $1\%, 2\%,5\%,10\%, 20\%$ of the total size of the training set. The loss of $0.33$ (lowest) is obtained at $2\%$. \ding{184} We examine various exploration strategies: no exploration, constant explorations at $2\%$ and $10\%$, and linearly and exponentially decaying explorations starting from $100\%$ to $5\%$ of the actions. We find that the linearly decaying strategy converges to the lowest loss per episode of $0.27$. \ding{185} Inspired by the Fixed Q-Targets strategy~\cite{mnih2013playing}, we reset the neural network weights and the optimizer state at various intervals during training ranging from no reset to every $1, 10, 50, \text{and}~100$ episodes. We find that resetting the network every $50$ episodes incurs the lowest loss of $0.34$. \ding{186} To examine the effect of various position initializations on training loss, we sample initial positions from various distributions (Gaussian, uniform) and using strategies similar to curriculum learning~\cite{bengio2009curriculum}, i.e., gradually increase the position sampling range 
with linear or exponential increments. 
The lower loss of $0.24$ on $\mathcal{N}(0, 0.02)$  compared to $0.28$ on  $\mathcal{N}(0, 0.09)$ confirms that initialization of position closer to the goal states increases the chances of success. However, the distributions $\mathcal{N}(0, 0.02)$ and $\mathcal{N}(0, 0.09)$ only sparsely encounter the full range of position initializations hence they are excluded from Figure~\ref{fig:sim_experiments}. The best strategy is determined to be exponentially increment the position using the idea of curriculum learning, which incurs a loss of $0.26$.

Overall, the result of training $300$ episodes in simulation using the best parameters and strategies found above is shown in Figure~\ref{fig:simulation} RIGHT. Among the $300$ episodes, the number of trajectories that successfully fulfill the control objective is $75$ with the average loss per episode in the last 10 steps equal to $0.17$ and an average training time of one minute and $53$ seconds (using a Macbook Air with an M1 processor and 8GB RAM).



%% file: sections/conclusion.tex
\section{Conclusion and Future Work}
\label{sec:conclusion}

In this work, we demonstrate that by using the Neural Fitted Q algorithm on a physical system, an RL agent can learn to fulfill a complex control objective based solely on real-time reward feedback without an offline training environment. Advantages of this method include short training time and inference time as well as minimal computational load because of a neural network with a small number of parameters and the sample efficiency of the algorithm. As a result, for a successful task completion at training, our approach uses only 10 minutes and 35 seconds in hardware and one minute and 53 seconds in simulation. Second, using the data collected from hardware interactions, we build a simulator to experiment a wide range of learning strategies.

This work can find its real-world applications in micromobility that deploys a modular autonomy approach, i.e., separated perception, prediction, and control components. After a decision is made on the desired steering angle, the DC motor can turn the steering wheel to achieve that angle via our approach. In addition, our approach can be used to tune system parameters according to the specific needs of a physical system based on factors such as load, size, and friction; a simulator such as the one shown in the work can be built from real-world data to facilitate this process.  

There are many future directions to pursue. On the system side, first, we would like to improve the convergence stability of our approach via testing the algorithm's limit on more complex hardware and control objectives w.r.t their desired time constraints. Second, we are interested in testing the approach's effectiveness and robustness on a physical system via computer vision~\cite{Shen2021Free}. Third, we want to develop a distributed algorithm to coordinate many physical systems that are separately controlled at different hardware complexities and time resolutions. We would also like to conduct stability analyses (after replacing the non-parametric KNN in simulation with a parameterized function) and sensitivity analyses of the controller for comparison with other traditional control methods. On the application side, we are interested in testing the approach in more complex simulated environments~\cite{Wilkie2015Virtual,Li2019ADAPS}, which can be calibrated to match real-world traffic conditions~\cite{Li2017CityFlowRecon,Li2017CityEstSparse,Lin2019ComSense}, provided more micromobility data is available.

\section{Acknowledgement}
This research is partially funded by NSF IIS 2153426. The authors would also like to thank NVIDIA and the University of Memphis for their support. 

%% file: main.bbl
\begin{thebibliography}{10}
\providecommand{\url}[1]{#1}
\csname url@samestyle\endcsname
\providecommand{\newblock}{\relax}
\providecommand{\bibinfo}[2]{#2}
\providecommand{\BIBentrySTDinterwordspacing}{\spaceskip=0pt\relax}
\providecommand{\BIBentryALTinterwordstretchfactor}{4}
\providecommand{\BIBentryALTinterwordspacing}{\spaceskip=\fontdimen2\font plus
\BIBentryALTinterwordstretchfactor\fontdimen3\font minus
  \fontdimen4\font\relax}
\providecommand{\BIBforeignlanguage}[2]{{%
\expandafter\ifx\csname l@#1\endcsname\relax
\typeout{** WARNING: IEEEtran.bst: No hyphenation pattern has been}%
\typeout{** loaded for the language `#1'. Using the pattern for}%
\typeout{** the default language instead.}%
\else
\language=\csname l@#1\endcsname
\fi
#2}}
\providecommand{\BIBdecl}{\relax}
\BIBdecl

\bibitem{cart2015}
L.~Hardesty, ``Self-driving golf carts,''
  \url{https://news.mit.edu/2015/autonomous-self-driving-golf-carts-0901},
  2015.

\bibitem{scooter1}
------, ``Driverless-vehicle options now include scooters,''
  \url{https://news.mit.edu/2016/driverless-scooters-1107}, 2016.

\bibitem{scooter2}
M.~McFarland, ``Self-driving scooters are coming to city sidewalks,''
  \url{https://www.cnn.com/2019/10/15/tech/self-driving-scooters-bikes/index.html},
  2019.

\bibitem{bike1}
D.~Silverberg, ``Mit is pioneering an `autonomous bicycle' and it's not just
  for lazy cyclists,''
  \url{www.vice.com/en/article/qvwxev/mit-is-pioneering-an-autonomous-bicycle-and-its-not-just-for-lazy]\\-cyclists},
  2018.

\bibitem{Lin2019BikeTRB}
L.~Lin, W.~Li, and S.~Peeta, ``Predicting station-level bike-sharing demands
  using graph convolutional neural network,'' in \emph{Transportation Research
  Board 98th Annual Meeting (TRB)}, 2019.

\bibitem{micro2021}
J.~Price, D.~Blackshear, W.~Blount, and L.~Sandt, ``Micromobility: A travel
  mode innovation,''
  \url{https://highways.dot.gov/public-roads/spring-2021/micromobility-travel-mode-innovation},
  2021.

\bibitem{mnih2013playing}
V.~Mnih, K.~Kavukcuoglu, D.~Silver, A.~Graves, I.~Antonoglou, D.~Wierstra, and
  M.~Riedmiller, ``Playing atari with deep reinforcement learning,''
  \emph{arXiv preprint arXiv:1312.5602}, 2013.

\bibitem{silver2016mastering}
D.~Silver, A.~Huang, C.~J. Maddison, A.~Guez, L.~Sifre, G.~Van Den~Driessche,
  J.~Schrittwieser, I.~Antonoglou, V.~Panneershelvam, M.~Lanctot \emph{et~al.},
  ``Mastering the game of go with deep neural networks and tree search,''
  \emph{nature}, vol. 529, no. 7587, pp. 484--489, 2016.

\bibitem{levine2016end}
S.~Levine, C.~Finn, T.~Darrell, and P.~Abbeel, ``End-to-end training of deep
  visuomotor policies,'' \emph{The Journal of Machine Learning Research},
  vol.~17, no.~1, pp. 1334--1373, 2016.

\bibitem{bahdanau2016actor}
D.~Bahdanau, P.~Brakel, K.~Xu, A.~Goyal, R.~Lowe, J.~Pineau, A.~Courville, and
  Y.~Bengio, ``An actor-critic algorithm for sequence prediction,'' \emph{arXiv
  preprint arXiv:1607.07086}, 2016.

\bibitem{el2013multiagent}
S.~El-Tantawy, B.~Abdulhai, and H.~Abdelgawad, ``Multiagent reinforcement
  learning for integrated network of adaptive traffic signal controllers
  (marlin-atsc): methodology and large-scale application on downtown toronto,''
  \emph{IEEE Transactions on Intelligent Transportation Systems}, vol.~14,
  no.~3, pp. 1140--1150, 2013.

\bibitem{wu2021flow}
C.~Wu, A.~R. Kreidieh, K.~Parvate, E.~Vinitsky, and A.~M. Bayen, ``Flow: A
  modular learning framework for mixed autonomy traffic,'' \emph{IEEE
  Transactions on Robotics}, 2021.

\bibitem{buechner2013nonlinear}
S.~Buechner, V.~Schreiber, A.~Amthor, C.~Ament, and M.~Eichhorn, ``Nonlinear
  modeling and identification of a dc-motor with friction and cogging,'' in
  \emph{IECON 2013-39th Annual Conference of the IEEE Industrial Electronics
  Society}.\hskip 1em plus 0.5em minus 0.4em\relax IEEE, 2013, pp. 3621--3627.

\bibitem{kara2004nonlinear}
T.~Kara and I.~Eker, ``Nonlinear modeling and identification of a dc motor for
  bidirectional operation with real time experiments,'' \emph{Energy Conversion
  and Management}, vol.~45, no. 7-8, pp. 1087--1106, 2004.

\bibitem{bellman1956bang}
R.~Bellman, I.~Glicksberg, and O.~Gross, ``On the “bang-bang” control
  problem,'' \emph{Quarterly of Applied Mathematics}, vol.~14, no.~1, pp.
  11--18, 1956.

\bibitem{aylor1980design}
J.~H. Aylor, R.~L. Ramey, and G.~Cook, ``Design and application of a
  microprocessor pid predictor controller,'' \emph{IEEE Transactions on
  Industrial Electronics and Control Instrumentation}, no.~3, pp. 133--137,
  1980.

\bibitem{qin2003survey}
S.~J. Qin and T.~A. Badgwell, ``A survey of industrial model predictive control
  technology,'' \emph{Control engineering practice}, vol.~11, no.~7, pp.
  733--764, 2003.

\bibitem{ross2011reduction}
S.~Ross, G.~Gordon, and D.~Bagnell, ``A reduction of imitation learning and
  structured prediction to no-regret online learning,'' in \emph{Proceedings of
  the Fourteenth International Conference on Artificial Intelligence and
  Statistics}, 2011, pp. 627--635.

\bibitem{sutton2018reinforcement}
R.~S. Sutton and A.~G. Barto, \emph{Reinforcement learning: An introduction},
  2018.

\bibitem{riedmiller2005neural}
M.~Riedmiller, ``Neural reinforcement learning to swing-up and balance a real
  pole,'' in \emph{2005 IEEE International Conference on Systems, Man and
  Cybernetics}, vol.~4.\hskip 1em plus 0.5em minus 0.4em\relax IEEE, 2005, pp.
  3191--3196.

\bibitem{hafner2007neural}
R.~Hafner and M.~Riedmiller, ``Neural reinforcement learning controllers for a
  real robot application,'' in \emph{Proceedings 2007 IEEE International
  Conference on Robotics and Automation}.\hskip 1em plus 0.5em minus
  0.4em\relax IEEE, 2007, pp. 2098--2103.

\bibitem{riedmiller2007learning}
M.~Riedmiller, M.~Montemerlo, and H.~Dahlkamp, ``Learning to drive a real car
  in 20 minutes,'' in \emph{2007 Frontiers in the Convergence of Bioscience and
  Information Technologies}.\hskip 1em plus 0.5em minus 0.4em\relax IEEE, 2007,
  pp. 645--650.

\bibitem{horng1999neural}
J.-H. Horng, ``Neural adaptive tracking control of a dc motor,''
  \emph{Information sciences}, vol. 118, no. 1-4, pp. 1--13, 1999.

\bibitem{meshram2012tuning}
P.~Meshram and R.~G. Kanojiya, ``Tuning of pid controller using ziegler-nichols
  method for speed control of dc motor,'' in \emph{IEEE-international
  conference on advances in engineering, science and management
  (ICAESM-2012)}.\hskip 1em plus 0.5em minus 0.4em\relax IEEE, 2012, pp.
  117--122.

\bibitem{thomas2009position}
N.~Thomas and D.~P. Poongodi, ``Position control of dc motor using genetic
  algorithm based pid controller,'' in \emph{Proceedings of the world congress
  on engineering}, vol.~2.\hskip 1em plus 0.5em minus 0.4em\relax London, UK,
  2009, pp. 1--3.

\bibitem{khater2015embedded}
A.~A. Khater, M.~El-Bardini, and N.~M. El-Rabaie, ``Embedded adaptive fuzzy
  controller based on reinforcement learning for dc motor with flexible
  shaft,'' \emph{Arabian Journal for Science and Engineering}, vol.~40, no.~8,
  pp. 2389--2406, 2015.

\bibitem{alkurawy2018model}
L.~E. Alkurawy and N.~Khamas, ``Model predictive control for dc motors,'' in
  \emph{2018 1st International Scientific Conference of Engineering
  Sciences-3rd Scientific Conference of Engineering Science (ISCES)}.\hskip 1em
  plus 0.5em minus 0.4em\relax IEEE, 2018, pp. 56--61.

\bibitem{syaichu2011model}
A.~Syaichu-Rohman and R.~Sirius, ``Model predictive control implementation on a
  programmable logic controller for dc motor speed control,'' in
  \emph{Proceedings of the 2011 International Conference on Electrical
  Engineering and Informatics}.\hskip 1em plus 0.5em minus 0.4em\relax IEEE,
  2011, pp. 1--4.

\bibitem{sahoo2015optimal}
S.~Sahoo, B.~Subudhi, and G.~Panda, ``Optimal speed control of dc motor using
  linear quadratic regulator and model predictive control,'' in \emph{2015
  international conference on energy, power and environment: towards
  sustainable growth (ICEPE)}.\hskip 1em plus 0.5em minus 0.4em\relax IEEE,
  2015, pp. 1--5.

\bibitem{lee2011model}
J.~H. Lee, ``Model predictive control: Review of the three decades of
  development,'' \emph{International Journal of Control, Automation and
  Systems}, vol.~9, no.~3, pp. 415--424, 2011.

\bibitem{vazquez2016model}
S.~Vazquez, J.~Rodriguez, M.~Rivera, L.~G. Franquelo, and M.~Norambuena,
  ``Model predictive control for power converters and drives: Advances and
  trends,'' \emph{IEEE Transactions on Industrial Electronics}, vol.~64, no.~2,
  pp. 935--947, 2016.

\bibitem{aamir2013replacing}
M.~Aamir, ``On replacing pid controller with ann controller for dc motor
  position control,'' \emph{arXiv preprint arXiv:1312.0148}, 2013.

\bibitem{caarls2015parallel}
W.~Caarls and E.~Schuitema, ``Parallel online temporal difference learning for
  motor control,'' \emph{IEEE transactions on neural networks and learning
  systems}, vol.~27, no.~7, pp. 1457--1468, 2015.

\bibitem{schenke2019controller}
M.~Schenke, W.~Kirchg{\"a}ssner, and O.~Wallscheid, ``Controller design for
  electrical drives by deep reinforcement learning: A proof of concept,''
  \emph{IEEE Transactions on Industrial Informatics}, vol.~16, no.~7, pp.
  4650--4658, 2019.

\bibitem{traue2020toward}
A.~Traue, G.~Book, W.~Kirchg{\"a}ssner, and O.~Wallscheid, ``Toward a
  reinforcement learning environment toolbox for intelligent electric motor
  control,'' \emph{IEEE Transactions on Neural Networks and Learning Systems},
  2020.

\bibitem{book2021transferring}
G.~Book, A.~Traue, P.~Balakrishna, A.~Brosch, M.~Schenke, S.~Hanke,
  W.~Kirchg{\"a}ssner, and O.~Wallscheid, ``Transferring online reinforcement
  learning for electric motor control from simulation to real-world
  experiments,'' \emph{IEEE Open Journal of Power Electronics}, vol.~2, pp.
  187--201, 2021.

\bibitem{watkins1989learning}
C.~J. C.~H. Watkins, ``Learning from delayed rewards,'' 1989.

\bibitem{schaul2015prioritized}
T.~Schaul, J.~Quan, I.~Antonoglou, and D.~Silver, ``Prioritized experience
  replay,'' \emph{arXiv preprint arXiv:1511.05952}, 2015.

\bibitem{van2016deep}
H.~Van~Hasselt, A.~Guez, and D.~Silver, ``Deep reinforcement learning with
  double q-learning,'' in \emph{Proceedings of the AAAI conference on
  artificial intelligence}, vol.~30, no.~1, 2016.

\bibitem{wang2016dueling}
Z.~Wang, T.~Schaul, M.~Hessel, H.~Hasselt, M.~Lanctot, and N.~Freitas,
  ``Dueling network architectures for deep reinforcement learning,'' in
  \emph{International conference on machine learning}.\hskip 1em plus 0.5em
  minus 0.4em\relax PMLR, 2016, pp. 1995--2003.

\bibitem{riedmiller201210}
M.~Riedmiller, ``10 steps and some tricks to set up neural reinforcement
  controllers,'' in \emph{Neural networks: tricks of the trade}.\hskip 1em plus
  0.5em minus 0.4em\relax Springer, 2012, pp. 735--757.

\bibitem{bengio2009curriculum}
Y.~Bengio, J.~Louradour, R.~Collobert, and J.~Weston, ``Curriculum learning,''
  in \emph{Proceedings of the 26th annual international conference on machine
  learning}, 2009, pp. 41--48.

\bibitem{Shen2021Free}
Y.~Shen, L.~Zheng, M.~Shu, W.~Li, T.~Goldstein, and M.~C. Lin, ``Gradient-free
  adversarial training against image corruption for learning-based steering,''
  in \emph{Thirty-fifth Conference on Neural Information Processing Systems
  (NeurIPS)}, 2021.

\bibitem{Wilkie2015Virtual}
D.~Wilkie, J.~Sewall, W.~Li, and M.~C. Lin, ``Virtualized traffic at
  metropolitan scales,'' \emph{Frontiers in Robotics and AI}, vol.~2, p.~11,
  2015.

\bibitem{Li2019ADAPS}
W.~Li, D.~Wolinski, and M.~C. Lin, ``{ADAPS}: Autonomous driving via principled
  simulations,'' in \emph{IEEE International Conference on Robotics and
  Automation (ICRA)}, 2019, pp. 7625--7631.

\bibitem{Li2017CityFlowRecon}
------, ``City-scale traffic animation using statistical learning and
  metamodel-based optimization,'' \emph{ACM Trans. Graph.}, vol.~36, no.~6, pp.
  200:1--200:12, Nov. 2017.

\bibitem{Li2017CityEstSparse}
W.~Li, D.~Nie, D.~Wilkie, and M.~C. Lin, ``Citywide estimation of traffic
  dynamics via sparse {GPS} traces,'' \emph{IEEE Intelligent Transportation
  Systems Magazine}, vol.~9, no.~3, pp. 100--113, 2017.

\bibitem{Lin2019ComSense}
L.~Lin, W.~Li, and S.~Peeta, ``Efficient data collection and accurate travel
  time estimation in a connected vehicle environment via real-time compressive
  sensing,'' \emph{Journal of Big Data Analytics in Transportation}, vol.~1,
  no.~2, pp. 95--107, 2019.

\end{thebibliography}
